\documentclass[letterpaper, 10 pt, conference]{ieeeconf} 

\IEEEoverridecommandlockouts                              
\usepackage{graphics} 
\usepackage{epsfig} 
\usepackage{times} 
\usepackage{amsmath} 
\usepackage{amssymb}  
\usepackage{romannum}

\usepackage{floatrow}
\usepackage{multirow}
\newcommand{\specialcell}[2][c]{%
\begin{tabular}[#1]{@{}c@{}}#2\end{tabular}}

\title{\LARGE \bf
EEG Classification based on Image Configuration in Social Anxiety Disorder
}

\author{\authorblockN{Lubna Shibly Mokatren$^{1}$, Rashid Ansari$^{1}$, Ahmet Enis Cetin$^{1}$\\
Alex D. Leow$^{2}$, Olusola Ajilore$^{2}$,  Heide Klumpp$^{2}$ and Fatos T.Yarman Vural $^{3}$}
\thanks{$^{1}$Department of Electrical and Computer Engineering,
        University of Illinois at Chicago, Chicago}%
\thanks{$^{2}$Department of Psychiatry, University of Illinois at Chicago, Chicago}%
\thanks{$^{3}$Department of Computer Engineering, Middle East Technical University, Ankara, Turkey
      }%
}

\begin{document}

\maketitle
\thispagestyle{empty}
\pagestyle{empty}

\begin{abstract}
The problem of detecting the presence of Social Anxiety Disorder (SAD) using Electroencephalography (EEG) for classification has seen limited study and is addressed with a new approach that seeks to exploit the knowledge of EEG sensor spatial configuration. Two classification models, one which ignores the configuration (model 1) and one that exploits it with different interpolation methods (model 2), are studied.  Performance of these two models is examined for analyzing 34 EEG data channels each consisting of five frequency bands and further decomposed with a filter bank. The data are collected from 64 subjects consisting of healthy controls and patients with SAD. Validity of our hypothesis that model 2 will significantly outperform model 1 is borne out in the results, with accuracy $6$--$7\%$ higher for model 2 for each machine learning algorithm we investigated. Convolutional Neural Networks (CNN) were found to provide much better performance than SVM and kNNs.
\end{abstract}

\section{Introduction}
Social Anxiety Disorder (SAD), world's third largest mental health care problem, affects $7\%$ of the population \cite{TheSocialAnxietyAssociation:2017:Online}. It is characterized by extreme fear and avoidance of social situations and the fear of negative evaluations from others \cite{leichsenring2017social}. The diagnosis process of SAD was first characterized in 1980 by Diagnosis and Statistical Manual for Mental Disorders (DSM-III). However, the criteria evolved and the most recent description appears in the fifth edition (DSM-5) \cite{hofmann2017cognitive}. In the field of psychiatry, there is the need to pay considerable attention to the reliability and quality of the diagnostic process of SAD DSM-5 to give an accurate assessment of the disorder \cite{kraemer2012dsm}. 

Electroencephalography (EEG) is a useful mechanism for diagnosing mental disorders. EEG provides measurements of brain activities aquired using electrodes placed over the scalp. While other brain imaging techniques, such as positron emission tomography (PET) and functional magnetic resonance imaging (fMRI), are used in diagnosis, EEG has important attributes in that it captures the temporal activity of the brain and is affordable compared with other methods \cite{sanei2013eeg}. The EEG waveform is usually divided into five main frequency bands \cite{article}: 
\\ Delta ($\delta$: up to 4 Hz) waves are generated during drowsiness and are the slowest, Theta ($\theta$: 4-8 Hz) waves are observed during quiet focus or sleep, Alpha ($\alpha$: 8 - 15 Hz) waves are observed during relaxation with closed eyes, Beta ($\beta$: 15-32 Hz) waves observed during normal consciousness and active thinking, and Gamma ($\gamma$: 32 Hz) associated with strong electrical signals caused by visual stimulation or information processing, learning, and perception. All the mentioned five frequency bands of the EEG signals are used in the analysis process for this study.

The use of EEG in SAD diagnosis has seen only limited study. The visual detection of differences in the EEG signals between SAD patient and control groups is impractical since the EEG activity appears to be similar in both. Therefore, automated detection using techniques such as machine learning is usually employed, which could lead to more precise diagnosis results and can be the first step for better connectivity analysis, pattern recognition process, and understanding treatment responses in SAD \cite{moscovitch2011frontal}.

In the past, many classification algorithms were devised for using EEG data \cite{lotte2007review}, such as, linear discriminant analysis, SVM, neural networks, nonlinear bayesian classifiers, kNN, hidden markov model, combination of classifiers, and others. However, none considered the spatial locations and configuration of the EEG channel sensors as a means to possibly achieving better accuracy in analysis or classification tasks.
This was a key driving factor in our research. Two classification models, one which ignores the configuration (model 1) and one that exploits it with different interpolation methods (model 2), are considered in this study. We hypothesized that model 2 will significantly outperform model 1. Validity of our hypothesis is borne out in the results, with average accuracy $7\%$ higher for model 2 for each machine learning algorithm we investigated. Convolutional Neural Networks (CNN) were found to provide much better performance than SVM and kNNs.

\section{METHOD}
\subsection{EEG recording}
The EEG data-set used in this paper was acquired in the Department of Psychiatry at University of Illinois at Chicago (UIC). The acquisition of multi-channel EEG is done using an Electro-Cap with electrodes positioned at 34 different locations i.e. channels. The data are gathered from a total of 64 subjects divided into control and SAD patients groups. For this study, the brain activity being analyzed is at resting state, without the introduction of stimuli or task instructions. The duration of the EEG recording varies from 2-7 minutes. The signals are sampled at 1024Hz sampling frequency.  

\subsection{Data Preprocessing}
EEG data are complex, as multiple processes take place simultaneously causing various artifacts and noise to appear, such as eye movements and EMG artifacts. Hence, it needs to be processed and cleaned for better interpolation during analysis. EEG data preprocessing stages are implemented using EEGLAB software. The frequencies of interest are in the range of 1-50 Hz, covering the five different frequency bands.

\subsection{Data Analysis}
Nonstationary phenomena are present in EEG data due to the constant switching of the meta-stable states of neurons assembling during brain functioning, causing signal changes in the form of spikes and momentary events. In our data analysis in time and frequency domains, each channel signal is divided into windows in which the data are assumed to be stationary. After testing multiple sizes of window segments based on the sampling frequency and the statistical properties, a window size of 5120 samples was found to yield the best results for detection.

The analysis of the signal content of each of the five main EEG frequency bands can be utilized to estimate subjects' cognition and emotional states. A dyadic wavelet packet transformation \cite{fliege1994multirate} is used for the decomposition of subbands corresponding to the five frequency bands. [0-4] Hz for $\delta$ band, [4-8] Hz for $\theta$, [8-16] Hz for $\alpha$, [16-32] Hz for $\beta$, and [32-52] Hz for $\gamma$.

The energy of the content of each windowed segment is computed for the five frequency bands separately. The analysis is based on the energy content of these signals represented in two different ways: concatenation of the channels of the five frequency bands and image-like 2D representation of the EEG channel locations. The latter method is discussed in Section \Romannum{2}.D. It should be noted that the outputs of the filter bank are ordered from highest frequency subband to the lowest.

\subsection{Image representation of the EEG data}
The data are acquired using 34 electrodes placed over different areas on the scalp, Frontal (F), Central (C) Temporal (T), Parietal (P) and Occipital (O) as shown in Fig. 1. It is hypothesized that the location of the channels can provide improved detection accuracy in the analysis of the data. To examine this hypothesis, two main data models are examined. First, the 34 channels of the five frequency bands are concatenated by creating a $34\times5$ energy matrix over each window, without accounting for the location of the channel electrodes. Second, a 3-D array of size $15\times15\times5$ is created, where the first two dimensions represent an image of $15\times15$ pixels corresponding with the channels positioning over the scalp while the third dimension represents the five frequency bands. In the latter method, the locations that not exactly correspond to any of the 34 channels are filled using different interpolation techniques.

The layout of the channels' location is given in Fig. 1. To construct an image-like representation of the electrodes layout, an image of size $15\times15$ was created. To fill in the missing pixel energy values, the following interpolation method was used \cite{eckstein1989evaluation}: an interpolated value $e$ at point $x$, only the samples $u_i=u(x_i)$ for $i=1,...n_i$, which lay within a distance less than $d_{max}$ from point $x$, are used to calculate interpolated value using the following weighted average.
\begin{equation}
u(x)=\frac{\sum_{i=1}^{n_i} w_i(x)u_i}{\sum_{i=1}^{n_i}w_i(x)}
\end{equation}
where $w_i(x)=\frac{1}{d(x,x_i)}$ and $d(x,x_i)$ represents the distance between points $x$ and $x_i$. This method is called Inverse Distance Weighting (IDW). In areas where no energy values exist within a distance of $d_{max}$ i.e. "Border Points" (BP) for abbreviation, the nearest value is simply repeated. $d_{max}$ is chosen empirically from a set of different distances.
\begin{figure}
\includegraphics[width=8cm]{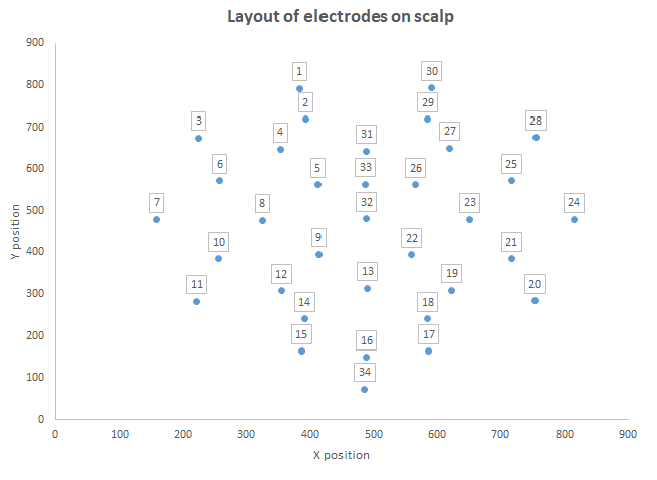}
\centering \caption{Layout of 34 electrodes on scalp}
 \label{fig1:Layout of 34 electrodes on scalp}
\end{figure}

It should be noted that other interpolation methods were examined such as nearest neighbor interpolation, bilinear interpolation, cubic spline interpolation \cite{lehmann1999survey}, and zero insertion at the border points (BP) with zero. However, they all result in inferior performance compared with the method mentioned above. Table \ref{table:4} in Section \Romannum{4} summarizes the average performance of the suggested methods. The construction of the image in the second model is tested on the main deep network structure (CNN) described in Section \Romannum{3}.B. 

\begin{figure*}[ht]
\scalebox{0.9}{\includegraphics[width=15cm]{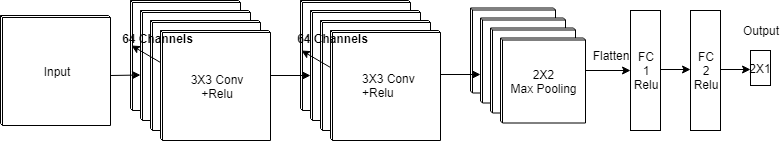}
\centering \caption{Convolutional neural network structure}
\hfill
 \label{fig2:Convolutional neural network structure}}
\end{figure*}

\section{EXPERIMENTS}
After the data are collected and preprocessed, two different models are used as previously discussed.
In the first model, the energy for the five frequency bands is calculated separately in each window of size 5120 samples. Hence, for each window an energy matrix of dimensions $34\times5$ is constructed. There are thirty four channels or electrodes and five frequency bands. Each row in the matrix represents one channel.
The second model is based on the image representation of the EEG data, as discussed in Section \Romannum{2}.D. For each window a 3D energy array of $15\times15\times5$ is built, where the third dimension represents the five frequency bands.
For both models, each matrix is considered as a single sample for the training or testing data-set.

\subsection{Acquisition of training and testing EEG data}
To train the network, a stratified 8-fold cross-validation is applied, the data are shuffled randomly and each fold is made by preserving the percentage of subjects from the two classes. 7 folds (56 subjects) are used for training and validation (about $10\%$ validation, $90\%$ training) and the remaining fold (8 subjects) is used for testing. The classifier is trained 8 separate times. In each trial a different fold is used for testing. Early stopping is applied by monitoring the validation loss to avoid overfitting.
In both models, for every window of size $N$, an energy matrix is constructed and considered as single sample, where $N=5120$. The samples are gathered for each of the training, validation and testing sets by sliding a moving window of size $N$ with no overlap of windows. The choice of sliding $N$ samples is made empirically as it yields better results when compared with other overlapping windows of shifts of $N/4$, $N/2$ and $3N/2$. In summary, the data set of each subject was divided into multiple samples, and the samples collected for the training, validation and testing sets will never overlap. Samples are labeled 0 (negative) if collected from control group, and 1 (positive) if collected from SAD patients.
The classification accuracy is the percentage of subjects classified correctly.

 For every trial, each testing subject is evaluated as follows: 
\begin{equation}
prediction=\left\{ \begin{array}{cl} patient & \mbox{if}  ~x_i=\frac{p_i}{t_i} \geq Th  \\ control & \mbox{else}  \end{array}\right.
\end{equation}
where  $Th=0.45$, $p_i$ is number of samples classified as positive for subject $i$, $t_i$ is total number of samples for subject $i$. Since some of the EEG data contain unwanted signals, the threshold was lowered from 0.5 to 0.45.

\subsection{CNN Network Structure}
 Deep learning has significantly improved the performance for many problems compared with other machine learning algorithms \cite{deng2014deep}. Developing deep network-based solutions has been found to be effective in many applications. CNNs have turned out to be the most powerful deep learning architectures for image-related problems \cite{krizhevsky2012imagenet}. The input to the network is arrays of data containing energy values of subband signals, in windows of size 5120.
A sequential model is built for this classification problem as shown in Fig. 2. The first layer is a 2D convolution with kernel size (convolution window) $3\times3$, 64 output filters, and ReLu activation over the outputs, another 2D convolution layer with the same parameters followed. This is followed by a max-pooling layer with pool size of 2, followed by dropout with rate=0.25. The input is then flattened which is followed by a fully connected layer with 128 output dimension and ReLu activation. Another dropout is done with rate=0.2 followed by final fully connected layer with Softmax activation and output dimension that equals to 2. Dropout is a regularization method that is used to reduce over-fitting. Other networks with one or more convolutional layers are also tested. However, all produced an inferior performance compared with the proposed network. The parameters of this network were chosen using several different trials and the parameters that yielded the best performance were selected. It should be noted that batch normalization was applied to the input and the CNN layers to reduce the internal covariate shift \cite{ioffe2015batch}. 

\section{RESULTS}
In this study, inputs constructed by the two main models are both tested, by feeding them to different machine learning algorithms or deep neural networks. Model 1 is built by concatenating the channels of the five frequency bands with no consideration for the spatial configuration of EEG electrodes. Model 2 takes the electrode configuration over the scalp into account. The two models are investigated for their classification performance using SVM, kNN, and a proposed CNN in Section \Romannum{3}.B. 

In the proposed CNN, the confusion matrices for Model 1 and Model 2 can be seen in Table \ref{Confusion Matrix 1} and  Table \ref{Confusion Matrix 2}, respectively. The confusion matrix summarizes the prediction results. The first entry of the first column has the number of actual positive (i.e., SAD patients) predicted as patients, the second entry of the first column has the number of actual patients predicted as negative (i.e., healthy subjects), etc.
The accuracy (ability to correctly classify a subject), sensitivity (ability to correctly identify patients when SAD is present), and specificity (ability to correctly identify control within a healthy group) are all higher for the proposed approach (Model 2), with overall classification accuracy of  $87\%$ for Model 2 and  $80\%$  for Model 1. Thus, Model 2 that takes advantage of the location of the channels is found to provide superior performance.

In this classification task the EEG input data are also fed to SVM with radial basis kernel function (RBF) with kernel parameter $\sigma=0.4$ using LIBSVM tool. The parameters were chosen by estimating the performance of SVM with different kernels and hyper-parameters tuning using the validation set.
The data are also applied to k-NN with k=3, and to other deep networks structures but the proposed network structure stood out as superior in terms of overall classification accuracy, as seen in Table \ref{table:3}. 
It should be noted that, in all cases, Model 2 gave significantly higher accuracy than Model 1, establishing the importance of 2D representation of the spatial configuration of EEG sensors.

\begin{table}[t]
\renewcommand{\arraystretch}{1.3}
\caption{Confusion Matrix - Model \Romannum{1}}
\label{Confusion Matrix 1}
\centering
\scalebox{0.8}{\begin{tabular}{|cc|c|c|l}
\cline{1-4} \bfseries Model \Romannum{1} 
& & \multicolumn{2}{ c| }{\bfseries Actual Condition} \\ \cline{3-4} 
& & Positive & Negative & Accuracy = $80\%$\\ \cline{1-4}
\multicolumn{1}{ |c  }{\multirow{2}{*}{\bfseries \specialcell[t]{Predicted\\Condition}}}  &
\multicolumn{1}{ ||c| }{Positive} & 26 & 7 & Sensitivity = $81\%$    \\ \cline{2-4}
\multicolumn{1}{ |c  }{}                        &
\multicolumn{1}{ ||c| }{Negitive} & 6 & 25 & Specifitty = $78\%$ \\ \cline{1-4}
\end{tabular}}
\end{table}

\begin{table}
\renewcommand{\arraystretch}{1.3}
\caption{Confusion Matrix - Model \Romannum{2}}
\label{Confusion Matrix 2}
\centering
\scalebox{0.8}{\begin{tabular}{|cc|c|c|l}
\cline{1-4} \bfseries Model \Romannum{2}  
& & \multicolumn{2}{ c| }{\bfseries Actual Condition} \\ \cline{3-4} 
& & Positive & Negative & Accuracy = $87\%$ \\ \cline{1-4}
\multicolumn{1}{ |c  }{\multirow{2}{*}{\bfseries \specialcell[t]{Predicted\\Condition}}}  &
\multicolumn{1}{ ||c| }{Positive} & 29 & 5 & Sensitivity = $90\%$    \\ \cline{2-4}
\multicolumn{1}{ |c  }{}                        &
\multicolumn{1}{ ||c| }{Negitive} & 3 & 27 &  Specificity = $84\%$ \\ \cline{1-4}
\end{tabular}}
\end{table}

\begin{table}[t]
\caption{Average classification performance}
\label{table:3}
\centering
\scalebox{0.8}{\begin{tabular}{ |p{3cm}||p{1.5cm}|p{1.5cm}|  }
 \hline
 \multicolumn{3}{|c|}{\bfseries Classification Accuracy} \\
 \hline
 Model     & Model \Romannum{1} &Model \Romannum{2}\\
 \hline
 Proposed CNN  &$80\%$ & $87\%$\\
 SVM   &$72\%$    &$79\%$\\
 kNN  &   $69\%$  & $75\%$ \\
 \hline
\end{tabular}}
\end{table}

To represent the different channels as an image over the scalp in the second model, a few interpolation methods are tested, as mentioned in Section \Romannum{2}.D. The average performance of these methods is presented in Table \ref{table:4}. Showing that Inverse Average Weighted interpolation yielded the best results among the methods tried.

\begin{table}[h!]
\centering
\scalebox{0.8}{\begin{tabular}{ |p{3cm}||p{4cm}| }
 \hline
 \multicolumn{2}{|c|}{\bfseries Interpolation Comparison} \\
 \hline
Interpolation method     &Average recognition perf.  \\
 \hline
 Nearest Neighbor  &   $73\%$ \\ 
 Bilinear &$75\%$    \\
  IDW with 0 BP  &$78\%$ \\
 Cubic spline   &$83\%$    \\
 IDW with NN BP  &$87\%$ \\
 \hline
\end{tabular}}
\caption{Average performance of different interpolation methods}
\label{table:4}
\end{table}

\section{CONCLUSION}
In this study, it is found experimentally that SAD patients can be identified based on their EEG signals. Furthermore, the spatial configuration of the EEG electrodes is used for the first time in a classification task. When taking advantage of the positions of the channels, the performance of the CNN network shows $7\%$ improvement in accuracy compared with the same CNN in which configuration is ignored.

\bibliographystyle{IEEEtran}
\bibliography{myRef}

\end{document}